%% file: main.tex
\documentclass[letterpaper]{article} 
\usepackage{aaai2026}  
\usepackage{times}  
\usepackage{helvet}  
\usepackage{courier}  
\usepackage[hyphens]{url}  
\usepackage{graphicx} 
\usepackage{natbib}  
\usepackage{caption} 
\usepackage{amsfonts}
\usepackage{algorithm}
\usepackage{algorithmic}

\usepackage{newfloat}
\usepackage{listings}
\DeclareCaptionStyle{ruled}{labelfont=normalfont,labelsep=colon,strut=off} 
\floatstyle{ruled}
\newfloat{listing}{tb}{lst}{}
\floatname{listing}{Listing}
\usepackage{dsfont}
\usepackage{amsmath}
\usepackage{xcolor} 
\usepackage{multirow}
\usepackage{svg}
\usepackage{subfiles}
\usepackage{tikz}
\usetikzlibrary{arrows.meta,shapes.geometric,calc}

\title{Simulation-Driven Railway Delay Prediction: An Imitation Learning Approach}
\author {
    Clément Elliker\textsuperscript{\rm 1},
    Jesse Read\textsuperscript{\rm 1},
    Sonia Vanier\textsuperscript{\rm 1},
    Albert Bifet\textsuperscript{\rm 2,3}
}
\affiliations {
    \textsuperscript{\rm 1}LIX (École Polytechnique, IP Paris, CNRS)\\
    \textsuperscript{\rm 2}AI Institute, University of Waikato\\
    \textsuperscript{\rm 3}LTCI, Télécom Paris, IP Paris \\
    clement.elliker@polytechnique.edu, jesse.read@polytechnique.edu, sonia.vanier@polytechnique.edu, albert.bifet@telecom-paris.fr
}

\usepackage{bibentry}

\long\def\BEGINOMIT#1\ENDOMIT{\relax}  

\begin{document}

\maketitle

\begin{abstract}
Reliable prediction of train delays is essential for enhancing the robustness and efficiency of railway transportation systems. In this work, we reframe delay forecasting as a stochastic simulation task, modeling state-transition dynamics through imitation learning. We introduce Drift-Corrected Imitation Learning (DCIL), a novel self-supervised algorithm that extends DAgger by incorporating distance-based drift correction, thereby mitigating covariate shift during rollouts without requiring access to an external oracle or adversarial schemes. Our approach synthesizes the dynamical fidelity of event-driven models with the representational capacity of data-driven methods, enabling uncertainty-aware forecasting via Monte Carlo simulation. We evaluate DCIL using a comprehensive real-world dataset from \textsc{Infrabel}, the Belgian railway infrastructure manager, which encompasses over three million train movements. Our results, focused on predictions up to 30 minutes ahead, demonstrate superior predictive performance of DCIL over traditional regression models and behavioral cloning on deep learning architectures, highlighting its effectiveness in capturing the sequential and uncertain nature of delay propagation in large-scale networks. 
\end{abstract}

\input{content.tex} 

\bibliography{references}

\input{appendix}

\end{document}

%% file: content.tex
\begin{links}
\link{Code}{https://github.com/orailix/rail-delay-simulator}
\end{links}

\section{Introduction}

Railway networks are vital infrastructure supporting sustainable, large-scale mobility globally, facilitating billions of passenger journeys annually. The extensive reliance on rail transport directly reflects service quality; hence, transport providers prioritize reliable, efficient, and user-friendly operations to fulfill passenger expectations. As a result, accurate delay prediction has become a critical research area, enabling commuters to anticipate disruptions and allowing operational personnel to proactively manage service impacts.

Following \cite{rossler2021discerning}, we distinguish primary delays, which stem from operational problems such as rolling stock failures, signal malfunctions, or severe weather, from secondary delays, the knock-on effects that propagate these initial disruptions as late-running trains interfere with downstream slot allocations. Due to the nature of accessible data, the delay prediction literature focuses mostly on modeling secondary delays, whose dynamics are governed by intricate spatiotemporal dependencies among interconnected services and infrastructure elements. 

As described in \cite{spanninger2022review}, these approaches can be grouped into two categories. The first, event-driven approaches, capture the interdependence between arrival and departure events using stochastic models (Graph Models \cite{goverde2010delay}, Markov Chains \cite{csahin2017markov}), offering interpretability, uncertainty quantification and modest data requirements.  The second, data-driven approaches, cast delay prediction as a supervised regression problem, allowing them to learn complex traffic dynamics from historical data using machine learning models (Linear Regression and Tree-based methods \cite{kecman2015predictive}, Neural Networks \cite{oneto2018train}, Transformers \cite{arthaud2024transformers}). Yet event-driven dynamical models insufficiently represent complex interactions, while data-driven one-shot regressors may undervalue the inherent temporal dependencies of successive events.

To leverage the advantages of both methodologies, we propose to frame delay prediction as a stochastic simulation problem: a policy is trained by imitation learning to reproduce the state-transition dynamics $p(s_{t+1}|s_t)$ observed in historical data. During roll-out, the policy stochastically predicts following states, enabling uncertainty quantification through Monte Carlo simulation. We combine the sequential nature of event-driven models with the representational power of data-driven methods to capture traffic complexity.

We propose \textsl{Drift-Corrected Imitation Learning } (DCIL), a self-supervised extension of DAgger's dataset-aggregation approach \cite{ross2011reduction}. Rather than relying on an expensive external oracle, DCIL applies distance-based drift correction during roll-outs by evaluating each candidate action’s induced next-state and choosing the one that minimizes a distance $\psi(\cdot,\cdot)$ back toward the expert’s subsequent state. This strategy mitigates covariate shift without resorting to complex adversarial \cite{ho2016generative} or inverse reinforcement learning \cite{ng2000algorithms} frameworks, outperforming Behavioral Cloning \cite{torabi2018behavioral}.

Our evaluation uses an extensive real-world dataset derived from publicly accessible operational logs from  \textsc{Infrabel}, the Belgian railway infrastructure manager. This dataset encompasses over three million train operations across three years, covering 682 stations and diverse service types (regional, intercity, high-speed) under various conditions (peak, off-peak, weekday, weekend, disruptions). We benchmark simulation‑based methods against regression across multiple architectures. Strikingly, we find that a 1.4-million-parameter Multi-Layer Perceptron trained with DCIL outperforms a 19-million-parameter Transformer trained with conventional regression on our task.

Our key contributions to delay forecasting are threefold:
\begin{enumerate}
\item We model delay prediction in a stochastic simulation framework, focusing learning on short-term dynamics;
\item We propose DCIL, a novel self-supervised imitation learning algorithm that effectively addresses covariate shift through drift correction; 
\item We conduct an extensive evaluation on a large-scale, real-world open dataset for predictions up to 30 minutes ahead, demonstrating substantial improvements in predictive accuracy over regression approaches.
\end{enumerate}


\section{Problem Definition}
\label{sec:probdef}

Let train \(i\) be present in the rail network at time \(t\). We define its feature vector \({s}_t^{(i)} \in \mathbb{R}^d\) where $d$ depends on the chosen feature-encoding configuration, and $s_t^{(i)}$ contains the following components:
\begin{itemize}
    \item train type;
    \item past and scheduled future stations and lines;
    \item station roles (\emph{departure}, \emph{arrival}, \emph{passage});
    \item theoretical passage times at stations;
    \item realised delays at past stations;
    \item temporal context (hour of day, day of week).
\end{itemize}

Stations are embedded using the first eight non-trivial eigenvectors of the normalised graph Laplacian of the rail-network graph \cite{belkin2003laplacian}, then per-node (row-wise) L2-normalised to yield eight-dimensional spectral coordinates that compactly encode topology-preserving neighborhoods. Line embeddings are obtained by averaging the embeddings of their constituent stations.

Let \(n\) denote the number of trains present at time \(t\). Then, we  define the full network \textbf{state} at time $t$ as 
\[
s_t \;=\; \bigl(s_t^{(i)}\bigr)_{i=1}^{n}
\]
i.e.\ all per‑train states at time $t$.

Let the \textbf{itinerary} of train \(i\) be defined as a sequence of $m$ stations $(l_1,\dots,l_m)$. We denote the \emph{scheduled} passage time at station $l_j$ as \(\tau_{l_j}^{(i)} \in \mathbb{R}\) and the \emph{actual} passage time by \(\hat{\tau}_{l_j}^{(i)} \in \mathbb{R}\). The resulting delay is
\[
  d_{l_j}^{(i)}
  \;=\;
  \hat{\tau}_{l_j}^{(i)} - \tau_{l_j}^{(i)},
  \qquad
  d_{l_j}^{(i)} \in \mathbb{R},
\]
so \(d_{l_j}^{(i)} > 0\) indicates lateness, \(d_{l_j}^{(i)} = 0\) on‑time operation, and \(d_{l_j}^{(i)} < 0\) earliness. For stations not yet visited, \(d_{l_j}^{(i)}\)  constitutes the quantity we want to predict.

Given the current network state \(s_t\), our goal is, for every train \(i\) and every future station \(l\), to model the conditional distribution
\begin{equation}
    \label{eq:1}
    p\!\left(d_{l}^{(i)} \,\middle|\, s_t\right),
\end{equation}
or, in a point‑prediction setting, some aspect of this distribution, such as its mean or median. 


\section{Background and Related Work}\label{sec:relwork}

\subsection{Regression for train delay prediction}

As noted in the introduction, train‑delay research divides into two broad streams: event‑driven approaches, which model the interdependence between arrival and departure events, and data‑driven approaches, which learn predictive models directly from observations \cite{spanninger2022review}.

Recent years have seen a surge in the latter, with studies exploring deep‑learning architectures—including Convolutional Neural Networks and bidirectional LSTMs \cite{guo2022cnn}, Graph Neural Networks \cite{heglund2020railway}, and Transformers \cite{arthaud2024transformers}. Although these models outperform simple baselines they are compared to, they are evaluated on private datasets, so results cannot be compared across studies.

To our knowledge, \cite{yang2024hybrid} presents the first large‑scale head‑to‑head comparison of data‑driven methods for train‑delay prediction. They find that transformer‑based architectures perform best on an open‑source dataset covering two Chinese high‑speed routes.

\paragraph{Datasets}
Only a handful of refereed papers rely on \emph{public} delay data. The best‑curated example is the \textsc{HSR‑Delay} corpus (3399 trains, 727 stations, Oct 2019–Jan 2020) released in \emph{Scientific Data} \cite{zhang2022high}. However, the dataset spans only four months on a limited number of trains, and the follow‑up study of \cite{yang2024hybrid} evaluates just two lines.
\cite{dekker2022modelling} model delay using diffusion on the Belgian open‑data feed (same dataset as our paper), but do not predict delays for individual trains; instead, they simulate the evolution of delay across clusters of stations over time.
\cite{lapamonpinyo2022real} scrape the U.S.\ Amtrak API for a single corridor.  
To our knowledge, no train delay prediction study yet covers an entire \emph{mixed‑traffic} national network over multiple years, leaving generalisation beyond high‑speed or single‑corridor settings largely untested.

\subsection{Simulation-based prediction}

Past work on railway network simulation relies on handcrafted microscopic, mesoscopic, and macroscopic simulators such as RailSys \cite{bendfeldt2000railsys}, OpenTrack \cite{nash2004railroad}, PETER \cite{koelemeijer2000peter}, or PROTON \cite{sipila2023simulations} that encode operating rules and dispatcher heuristics instead of learning from historical data. Macroscopic simulators neglect precise train dynamics but scale well to large networks, whereas microscopic ones demand extensive calibration and significant compute, and mesoscopic models target only critical sections \cite{tiong2023quantitative}. Consequently, forecasts from these rule‑driven tools often fail to match real‑world delay patterns, especially on busy networks.

The work on rail-network simulation reflects that of the wider time-series forecasting literature. Train delay is deterministically linked to train trajectory (i.e., delay at $t$ can be derived from state $s_t$), with trajectory prediction being a specific case of time-series forecasting (cf. Eq.~\eqref{eq:1}). 


A traditional approach to time series forecasting, often used in the context of signal processing, is to implement a dynamical model $p(s_{t+1} \mid s_t)$ whose parameters have a physical meaning (speed, acceleration, etc.), e.g. \cite{prevost2007extended,TSPF}. 

Specifying reliable dynamical models with an expert is not always feasible (particularly with complicating external factors such as human behaviour involved), but such models can instead be learned in a data-driven (and often, model-agnostic) approach via machine-learning; essentially learning $p(s_{t+1} \mid s_t)$ or similar from training pairs $\{(s_{t}, s_{t+1})\}_t$. 


Data‑driven simulation approaches have shown their efficiency in predicting future events in various domains. In weather forecasting, \cite{price2025probabilistic} introduced GenCast, a conditional diffusion model trained on decades of reanalysis data that generates 15‑day global ensemble forecasts in minutes and surpasses the leading physics‑based system (ECMWF‑ENS) on 97\% of evaluation targets. In road‑traffic modeling, \cite{kuefler2017imitating} applied Generative Adversarial Imitation Learning to train a driver‑behavior simulator that reproduces realistic lane changes, speed profiles and collision‑free trajectories, showing that imitation‑learning‑based simulation can successfully predict future traffic events.

\subsection{Imitation Learning}

Imitation learning seeks a policy $\pi_\theta$ that reproduces expert behaviour from demonstration trajectories $\tau=(s_0,a_0,\dots,s_T,a_T)$. Four families dominate the literature:

\noindent\textbf{Behavioral Cloning (BC).} Treats imitation as supervised learning: fits $\pi_\theta(s)$ to expert actions $a$ on recorded $(s,a)$ pairs. Fast and data‑efficient but prone to covariate shift once the learner visits states absent from the dataset.

\noindent\textbf{Dataset Aggregation (DAgger).} Iteratively executes the current policy, queries the expert on the visited states, and augments the training set. This feedback loop curbs covariate shift but can be expensive when expert labels are costly.

\noindent\textbf{Inverse Reinforcement Learning (IRL).}  Alternates between (i) fitting a reward function under which the expert is (near‑)optimal and (ii) training the policy via RL using learned rewards. This bi‑level loop generalises beyond the demonstration manifold but is computationally heavy.

\noindent\textbf{Generative Adversarial Imitation Learning (GAIL).}  Derived from IRL, GAIL sets up a GAN‑style game: a discriminator distinguishes expert from learner state–action pairs, and its negative log‑probability serves as the reward fed to the policy optimiser (e.g.\ TRPO/PPO).  This removes the expensive bi-level loop, but inherits typical GAN instabilities such as mode collapse.


\section{Delay Prediction via Macroscopic Simulation}
\label{sec:sim}

\begin{figure}[t]
\centering
\begin{tikzpicture}[
    xscale=1.8,          
    yscale=2.5,
    >=Stealth,
    station/.style={circle,draw,fill=blue!60,minimum size=10pt,inner sep=0pt},
    train/.style={diamond,draw,fill=red!70,minimum size=8pt,inner sep=1.5pt},
    font=\small
]

\draw (-1.5,0) -- (1.5,0);

\node[station] (s1) at (-1,0) {};
\node[station] (s2) at (0,0) {};
\node[station] (s3) at (1,0) {};

\node[train] (tr) at (-0.5,0) {};

\coordinate (ctrA) at ([yshift=6pt]tr.center);

\draw[<-,>=Stealth,scale=1,>={Stealth[length=2pt,width=3pt]}]
  (ctrA) ++(305:0.1)              
    arc (305:595:0.1)            
    node[midway,above,yshift=1pt] {$a = 0$};

\coordinate (ctrB) at ([yshift=-6pt]tr.center);
\coordinate (ctrC) at ([yshift=-6pt]$(s2)!0.5!(s3)$);

\draw[->,bend left=-30]
  (ctrB) to node[below,yshift=-2pt] {$a = 1$} (ctrC);

\node[station] (legS) at (2.1,0.4) {};
\node[anchor=west] at (2.2,0.4) {stations};

\node[train] (legT) at (2.1,0.15) {};
\node[anchor=west] at (2.2,0.15) {train};

\end{tikzpicture}
\caption{Illustration of actions}
\label{fig:actions_illus}
\end{figure}
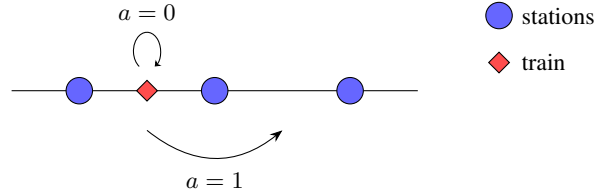

We frame the railway network as a Markov Decision Process (MDP). Let $s_t = (s_t^{(1)}, ..., s_t^{(n)})$ be the state of the rail network at time $t$, defined in Section~\ref{sec:probdef}.  Let \( \mathcal{A} \) denote the action space, with $a_t = (a_t^{(1)}, ..., a_t^{(n)})$ the actions describing the movements of trains on the network at time $t$. As illustrated in Figure~\ref{fig:actions_illus}, action $a^{(i)}\in\{0,1,2\}$ moves train $i$ by that many scheduled stations during the next time step, with the cap at 2 ensuring a finite discrete action space. In this work, we set $\Delta t = 30$ seconds, so $s_{t+1}$ represents a full \emph{snapshot} of the network 30 seconds after $s_t$. No hand-crafted block-constraint rules are enforced; instead, the policy approximates them implicitly by imitation, as a precise specification is impractical. Actions and states remain discrete to fit available data, a continuous version would require GPS.

The environment dynamics, applied per train, are defined by the function $\phi:\mathcal{S}^{(i)} \times \mathcal{A}^{(i)} \rightarrow  \mathcal{S}^{(i)}$, so that \ $s_{t+1}^{(i)} = \phi(s_t^{(i)},a_t^{(i)})$. For simplicity, we still denote the transformation as $\phi(s, a)$, mapping a set of train-action pairs $(s, a)=((s^{(1)}, ..., s^{(n)}), (a^{(1)}, ..., a^{(n)}))$ to their next state $(\phi(s^{(1)}, a^{(1)}), ..., \phi(s^{(n)}, a^{(n)}))$.
As a result, we obtain
 \[
p_{\pi}(s'|s) = \sum_{a \in \mathcal{A}} \pi(a|s) \mathds{1}[\phi(s,a) = s'] 
 \]
 whose sum has at most one non‑zero term since $\phi$ is injective for a fixed $s$: by construction, distinct actions map to distinct states, with $\pi(a|s)=\displaystyle \prod_{i=1}^n \pi_i(a^{(i)}|s)$.
 
As delay can be deduced from states, we rewrite delay prediction as prediction of future states:

\[
p_{\pi}(d_l^{(i)} \,|\, s_0) = f(p_{\pi}(s_1,\dots,s_T\,|\,s_0))
\]

where $f$ retrieves the delay from the predicted states. 

Via Markov assumption:
\[
p_{\pi}(s_1,\ldots,s_T|s_0) = \prod_{t=1}^{T} p_{\pi}(s_{t}|s_{t-1})
\]

Exact evaluation is infeasible, so we approximate the distribution with
Monte Carlo rollouts of policy $\pi$. In this work, we obtain the delay point forecast by retrieving the median of the empirical distribution from the sampled trajectories.




In summary, we have reduced delay prediction to learning a policy $\pi$ that approximates $p(a|s)$. The following Section presents the process of learning said policy. 

\section{Drift-Corrected Imitation Learning (DCIL)}\label{sec:dcil}

We assume an (implicit) expert policy $\pi_*$ has generated a logged dataset of train trajectories,
\[
\mathcal D=\bigl\{\left((s_{0},a_{0}),\dots,(s_{T},a_{T})\right)_n\bigr\}_{n=1}^{|\mathcal{D}|}
\qquad
a_t \sim \pi_*(\cdot \mid s_t).
\]

The term “expert” is purely notional: it refers to the behavioural patterns encoded in historical train‑movement data rather than to a conscious decision‑maker.  Each recorded trajectory is treated as if it were sampled from $\pi_*$.

Our goal is to learn a policy $\pi_{\theta}$ via imitation learning that closely matches the expert action distribution $\pi_*(a_t \mid s_t)$.

The standard method within Imitation Learning is Behavioural Cloning (BC): it trains a parametric policy $\pi_{\theta}$ by minimising the cross‐entropy between the predicted action distribution $\pi_{\theta}(\cdot | s)$ and the one‑hot expert labels from the demonstration set $\mathcal{D}$.
However, this method is notoriously prone to covariate shift: prediction errors push the policy into unseen states, where its performance quickly degrades. In order to overcome this limitation in our setting, we propose Drift‑Corrected Imitation Learning (DCIL), a self‑supervised extension of DAgger that injects simulator‑generated corrective labels during training, counteracting covariate shift without requiring any additional expert queries or complex adversarial schemes.

Starting from a state \(s_0\) drawn from a demonstration
trajectory $(s_0,\dots,s_T) \in \mathcal D$, we roll out our policy \(\pi_\theta\)
for \(T\) steps to obtain a \emph{policy trajectory}
\((s_0,s_1',\dots,s_T')\).  For each intermediate state \(s_t'\) we
construct a synthetic “expert” action \(a_t^\ast\) by selecting, for each train $i$ in the network, the action $a^{\ast(i)}_{t}$ that brings it closest to its next ground truth state $s^{\ast(i)}_{t+1}$:
\[
  a_t^{\ast(i)} \;=\;
  \arg\min_{a\in\mathcal{A}^{(i)}}
         \psi\!\bigl(s_{t+1}^{(i)},\, \phi(s_t'^{(i)},a)\bigr),
\]
where \(\psi:\mathcal S^{(i)}\times\mathcal S^{(i)}\!\to\!\mathbb R_{>0}\) is a  distance in state space. In this work, $\psi$ computes the number of stations separating the two individual states $s^{(i)}$ and $s^{'(i)}$ along the itinerary of train $i$. At each step $t$, we choose for each train the action that minimizes the \emph{future} distance, i.e., the action $a^{*(i)}$ is 0 when the train is at the right station or ahead, 1 when it is one station behind, and 2 when it is 2 or more stations behind. 

This algorithm is off-policy. As a result, we store synthetic state-action pairs in a replay buffer and use them for multiple epochs to improve sample efficiency.

Because synthetic labels become less reliable as the rollout drifts
farther from the expert trajectory, we down‑weight their influence
according to the distance between the current step $t$ policy states $s_t'^{(i)}$ and the next expert state $s_{t+1}^{(i)}$. One SGD step on the cross‑entropy loss uses the scaled gradient
\begin{equation}\label{eq:dcilstep}
    \sum_i
    \frac{1}{1+\alpha\, \psi(s_{t+1}^{(i)},s_t'^{(i)})^{\beta}}\;
  \nabla_\theta\,
  \ell\!\bigl(\pi_\theta(s_t')^{(i)},\,a_t^{\ast(i)}\bigr)  
\end{equation}

with hyperparameters \(\alpha>0\) and \(\beta\ge 1\).  Thus, trajectories very close to the expert receive nearly full weight, while poorly matching ones contribute only marginally.

The full training loop, described in Algorithm~\ref{alg:dcil}, starts with initializing $\pi_{\theta}$ and $B$. Then, for $E$ epochs, it (i) rolls out the current policy to inject \(n_s\) synthetic samples, discarding the oldest to keep the buffer at capacity~\(C\), and (ii) performs Adam updates on mini‑batches drawn from the entire buffer $B$, where each gradient is down‑weighted according to the distance‑based weight in Eq.\,\eqref{eq:dcilstep}.

\begin{algorithm}[t]
\caption{Drift-Corrected Imitation Learning (DCIL)}
\label{alg:dcil}
\begin{algorithmic}[1]
\REQUIRE demonstrations $\mathcal D$; horizon $T$;
         buffer capacity $C$; new samples per epoch $n_s$;
         epochs $E$; mini-batch size $m$; learning rate $\eta$;
         weighting factors $\alpha,\beta$
\STATE initialise policy parameters $\theta$
\STATE initialise empty replay buffer $B$
\FOR{$e = 1$ \textbf{to} $E$}
  \STATE $k \leftarrow 0$
  \WHILE{$k < n_s$}
    \STATE sample $(s_0,\ldots,s_T) \sim \mathcal D$
    \STATE roll out $(s_0,s_1',\ldots,s_T') \sim p_{\pi_\theta}$ from $s_0$
    \FOR{$t = 0,\ldots,T-1$ \textbf{and} $k < n_s$}
      \STATE $a_t^\ast$: $a_t^{\ast(i)}=\arg\min_{a\in\mathcal A^{(i)}}\psi(s_{t+1}^{(i)},\phi(s_t'^{(i)},a))$
      \STATE $w_t: w_t^{(i)}=\tfrac{1}{1+\alpha\,\psi(s_{t+1}^{(i)},s_t'^{(i)})^{\beta}}$
      \STATE push $(s_t',a_t^\ast,w_t)$ into $B$; \textbf{if} $|B|>C$ \textbf{then} discard oldest
      \STATE $k \leftarrow k+1$
    \ENDFOR
  \ENDWHILE
  \FOR{\text{each mini-batch } $M \subset B$ of size $m$}
    \STATE $g \leftarrow \nabla_\theta \displaystyle \sum_{(s',a^\ast,w)\in M} \sum_i w^{(i)}\,\ell(\pi_\theta(s')^{(i)},a^{\ast(i)})$
    \STATE $\theta \leftarrow \text{Adam}(\theta,g,\eta)$
  \ENDFOR
\ENDFOR
\end{algorithmic}
\end{algorithm}

\section{Experimental Evaluation}\label{sec:experiments}

\subsection{Data}
Our empirical evaluation uses raw operational logs provided by \textsc{Infrabel}, the Belgian railway infrastructure manager, covering a three-year period from \textbf{1 January 2022} to \textbf{31 December 2024}.
 Every 30 seconds, we build a network‑wide \emph{snapshot} that contains the vector encoding of each train, as per Section~\ref{sec:probdef}.  Network density varies from fewer than 10 trains during late‑night service to more than 400 at peak hour. To augment the schedule context, we insert each train at a \emph{placeholder} station 5~min before its planned departure and keep it at a \emph{placeholder} station for 5~min after its final observed stop.  Records with missing or incoherent timestamps are discarded (less than 1\% of trains).  Finally, we uniformly subsample 10\% of the \emph{snapshots}, yielding \textbf{255\,k} \emph{snapshots} with \textbf{51 million} train instances in total.

We adopt a strict temporal split to avoid information leakage and match industrial settings with a full calendar year for testing to maximise seasonal and operational diversity:
\begin{itemize}
    \item \textbf{Training}: 2022‑01‑01 $\rightarrow$ 2023‑09‑30
    \item \textbf{Validation}: 2023‑10‑01 $\rightarrow$ 2023‑12‑31
    \item \textbf{Test}: 2024‑01‑01 $\rightarrow$ 2024‑12‑31
\end{itemize}

Evaluating a single \emph{snapshot} with the simulation‐based approach requires sampling 50 trajectories and simulating 66 time steps; on an NVIDIA A100 this takes about 1.5 seconds of wall‑clock time.  To keep evaluation tractable, we down‑sample the test split to 800 \emph{snapshots} ($\approx$ 200 trains per \emph{snapshot}), yielding \textbf{1.6 million} arrival‑time predictions ($\approx$ 10 per train) that are produced in roughly 20 min end‑to‑end. Test snapshots are uniformly drawn, but busy periods produce more predictions, effectively emphasizing peaks.

\emph{Reproducibility.} All experiments are fully reproducible; code and appendix are available on GitHub (link on the first page).

\subsection{Method-Specific Details}

In this work, we are interested in multi-station delay forecasting: i.e., predicting delay for the next $n$ stations. This section provides implementation details for the evaluated methods (regression, BC, DCIL) and architectures (XGBoost, MLP, Transformer).

\subsubsection{Regression target}

Following Section~\ref{sec:probdef}, we model delay via the conditional
distribution \(p\bigl(d^{(i)}_{l}\,\bigl|\,s_t\bigr)\). For a train \(i\) travelling on itinerary \(l\) and currently located at station index \(j\), we need a forecast for the next $n$ stations. 
Rather than predicting absolute delays, we regress on the difference to the last known delay
\[
\Delta d^{(i)}_{l_{j+1}:l_{j+n}}
  :=\bigl(d^{(i)}_{l_{j+1}}-d^{(i)}_{l_j},\,\dots,\,
          d^{(i)}_{l_{j+n}}-d^{(i)}_{l_j}\bigr).
\]
Our model outputs a point estimate $\Delta \hat{d}^{(i)}_{l_{j+1}:l_{j+n}}$ conditionally on $s_t$; we then recover absolute delays by adding the last known delay \(d^{(i)}_{l_j}\), yielding the prediction $\hat{d}^{(i)}_{l_{j+1}:l_{j+n}}$.

\subsubsection{Ensuring fair comparison}
To guarantee that regression and simulation models are evaluated on \textbf{exactly the same targets}, we restrict the predictive horizon to \(H = 30\text{ min}\).  In the final metrics, we therefore retain only station events whose observed arrival time lies within \(H\) of the reference \emph{snapshot} \(s_0\).  Because the regression baselines output a fixed‑length vector, we set the output length to the next \(K = 15\) stations—a compromise that covers almost all instances within 30 min while avoiding an unnecessarily large head that would penalise regression.  The simulator is forced to respect the same limit and produces at most \(K\) predictions per train.  Should a simulated train fail to reach one of the \(K\) future stations within the simulated steps (e.g.\ it keeps choosing \textsc{stay}: $\text{a}=0$), we copy forward its most recent known delay and, if necessary, enlarge it so that the predicted arrival time never precedes the last simulated time step.  To allow the simulator to make mistakes by predicting arrival times beyond the predictive horizon, we simulate 10\% more time steps.

\subsubsection{Method‑dependent station/line horizon}
The raw features of Section~\ref{sec:probdef} are computed for the previous five visited stations and lines for \emph{all} models. To capture information about upcoming stations and lines, we include the next five scheduled stations/lines when the input feeds the simulator and the next fifteen for a regression model.

\paragraph{Mitigating Stalled Policies in BC}
During roll‑out, we observed that the BC learned policy occasionally \emph{stalls}: it assigns almost all probability mass to the \textsc{stay} action (\(a = 0\)) at every step, so the train never advances. We keep for each train a running floor \(\mu\), defined as the largest value of \(\pi(a{=}1)\) since that train last advanced.  At every step, we clamp \(\pi(a{=}1) \leftarrow \max\!\bigl(\pi(a{=}1),\,\mu\bigr)\) and then renormalise the action distribution; this simple trick reduces the fraction of stalled trains and improves long‑horizon accuracy. This isn't necessary for DCIL.

\subsection{Architecture‑Specific Details}
\label{subsec:arch}

To highlight the contrast between \emph{regression} and \emph{simulation} paradigms, we benchmark three architectures of ascending expressive power: a classical gradient‑boosted tree ensemble (\textbf{XGBoost}), a lightweight \textbf{multi‑layer perceptron} (\textbf{MLP}), and an encoder‑only \textbf{Transformer}.

\paragraph{Transformer.}
Following \cite{arthaud2024transformers}, each input token represents one train, i.e.\  $s_t^{(i)}$ defined in Section~\ref{sec:probdef}. Through the attention mechanism, the model can propagate information between trains, predicting: (i) delay for regression and (ii) probability over the action space for simulation, conditionally to the state of the network.

\subsubsection{MLP/XGBoost}

Due to the fixed‐length input nature of both the multi‑layer perceptron (MLP) and XGBoost, we cannot feed them the variable‑size set of trains present in a network \emph{snapshot} the way the Transformer does. As a result, the model inputs are train-specific rather than snapshot-specific.

In order to give the model network-related information, we build features that we concatenate with those introduced in Section~\ref{sec:probdef}. For \emph{each} train, we compute, at the five radii \(r\!\in\!\{0.1, 0.3, 0.6, 1.0, 2.0\}\) (chosen ad hoc to capture different ranges of network interactions), 
\begin{enumerate}
    \item the \emph{count} of neighbouring trains lying within Euclidean distance~\(r\) in the embedding space, min-max normalised
    \item the \emph{mean} of the past delays of trains in the neighborhood
\end{enumerate}
These ten scalars provide a compact summary of the local traffic context; appending them to the base feature vector yields the fixed‑length representation required by both the MLP and XGBoost models.


\paragraph{Hyperparameter tuning}

Table~\ref{tab:gen} and Table~\ref{tab:hor} are produced with the following two‑phase protocol.  
First, for each architecture we carry out a \emph{model‑specific, multi‑stage grid search} over a subset of hyper‑parameters, using mean absolute error (MAE) on the validation set as the selection metric.  
Second, the best configuration is \textbf{run} ten times with different random seeds; each run is retrained on the combined train and validation sets and evaluated on the held‑out test set. These runs are compiled using mean ± standard deviation across metrics.

\begin{table}[t]
\centering
\small
\renewcommand{\arraystretch}{1.2}
\begin{tabular}{llll}
\cline{1-4}
 Model & Method  & MAE & RMSE \\
\cline{1-4}
\multirow[t]{3}{*}{Transformer} & Regression & 60.53 ± 0.33 & 116.21 ± 4.06 \\
 & BC & 56.17 ± 1.02 & 106.78 ± 2.82 \\
 & DCIL & \textbf{52.24 ± 0.31} & \textbf{96.34 ± 0.58} \\
\cline{1-4}
\multirow[t]{3}{*}{MLP} & Regression & 64.34 ± 0.84 & 112.20 ± 2.37 \\
 & BC & 59.10 ± 0.67 & 107.29 ± 1.39 \\
 & DCIL & \textbf{57.76 ± 0.41} & \textbf{105.23 ± 0.73} \\
\cline{1-4}
\multirow[t]{2}{*}{XGBoost} & Regression & \textbf{64.48 ± 0.02} & \textbf{117.59 ± 0.19} \\
 & BC & 66.77 ± 0.04 & 120.42 ± 0.19 \\
\cline{1-4}
\end{tabular}
\caption{Test-set errors (mean ± std), all predictions.}
\label{tab:gen}
\end{table}

\begin{table*}[t]
\centering
\small
\renewcommand{\arraystretch}{1.2}
\begin{tabular}{llllllll}
\cline{1-8}
Model & Method & $\text{MAE}_{0-5}$ & $\text{MAE}_{5-10}$ & $\text{MAE}_{10-15}$ & $\text{MAE}_{15-20}$ & $\text{MAE}_{20-25}$ & $\text{MAE}_{25-30}$ \\
\cline{1-8}
\multirow[t]{3}{*}{Transformer} & Regression & 30.63 ± 0.22 & 45.42 ± 0.27 & 58.67 ± 0.29 & 70.25 ± 0.42 & 80.27 ± 0.46 & 90.35 ± 0.53 \\
 & BC & 24.14 ± 0.50 & 40.24 ± 0.86 & 54.39 ± 1.12 & 66.51 ± 1.28 & 77.17 ± 1.42 & 87.75 ± 1.30 \\
 & DCIL & \textbf{22.31 ± 0.30} & \textbf{36.88 ± 0.40} & \textbf{50.02 ± 0.43} & \textbf{61.53 ± 0.44} & \textbf{72.11 ± 0.37} & \textbf{83.38 ± 0.51} \\
\cline{1-8}
\multirow[t]{3}{*}{MLP} & Regression & 32.54 ± 0.87 & 48.48 ± 0.79 & 62.64 ± 0.94 & 74.73 ± 1.01 & 85.28 ± 1.02 & 95.40 ± 0.82 \\
 & BC & 24.93 ± 0.52 & 42.36 ± 0.75 & 57.24 ± 0.90 & 70.04 ± 0.91 & 81.29 ± 0.94 & 92.71 ± 1.31 \\
 & DCIL & \textbf{23.98 ± 0.42} & \textbf{41.06 ± 0.62} & \textbf{55.86 ± 0.65} & \textbf{68.55 ± 0.55} & \textbf{79.73 ± 0.37} & \textbf{91.29 ± 0.54} \\
\cline{1-8}
\multirow[t]{2}{*}{XGBoost} & Regression & 31.31 ± 0.03 & 48.13 ± 0.03 & \textbf{62.84 ± 0.04} & \textbf{75.75 ± 0.03} & \textbf{86.25 ± 0.03} & \textbf{95.94 ± 0.03} \\
 & BC & \textbf{27.95 ± 0.04} & \textbf{47.65 ± 0.06} & 64.38 ± 0.09 & 79.10 ± 0.07 & 92.28 ± 0.08 & 105.31 ± 0.09 \\
\cline{1-8}
\end{tabular}
\caption{Test-set MAE (mean ± std) for different predictive horizons (5-minute bins)}
\label{tab:hor}
\end{table*}

\section{Results and Discussion}
\label{sec:results}

Table~\ref{tab:gen} reports the mean and standard deviations of test‑set errors from 10 runs for all three model families—Transformer, MLP and XGBoost—under the three training methods: \emph{Regression}, \emph{Behavioural Cloning} (BC, the imitation‑learning baseline), and \emph{Drift‑Corrected Imitation Learning} (DCIL, our proposed imitation‑learning method). BC and DCIL predictions are retrieved using the median of an ensemble of 50 sampled trajectories. Since we rescale predictions and labels back to their original units, evaluation metrics such as Mean Absolute Error (MAE) and Root Mean Squared Error (RMSE) are measured in seconds. Lower values indicate better performance for both metrics.

For Transformer, moving from pure Regression to the simulation baseline BC already yields a clear improvement: MAE drops by \(7.2\%\) and RMSE by \(8.1\%\).
DCIL amplifies these gains, ultimately reducing MAE by \(13.7\%\) and RMSE by \(17.1\%\), respectively, relative to Regression, and by roughly half of that again relative to BC. The accompanying lower standard deviations hint at a stabler optimisation compared to BC when trajectory‑level feedback is used.
Regression yields surprisingly worse results for RMSE, hinting at a lack of robustness to outliers.

The same pattern holds for the MLP.  
BC narrows the gap to Regression (‑8.1\% MAE and ‑4.4\% RMSE), while DCIL delivers the best absolute scores and the lowest variance (‑10.2\% MAE and ‑6.2\% RMSE versus Regression). Thus, even a lightweight neural network benefits from simulation. Noticeably, the gap between BC and DCIL is smaller.

In contrast, the tree‑based XGBoost does not profit from the imitation‑learning signal. Switching from Regression to BC actually \emph{increases} the errors by 3–5\% across the board. DCIL was not evaluated here because the iterative training scheme is not compatible with XGBoost.

With DCIL, the Transformer achieves the lowest overall errors (MAE\,=\,52.24, RMSE\,=\,96.34).
Interestingly, even the simpler MLP architecture (1.4M parameters), when trained with imitation learning, surpasses the Transformer (19M parameters) trained purely by regression: MLP‑BC attains an MAE of 59.10 s and an RMSE of 107.29 s, while MLP‑DCIL improves further to 57.76 s and 105.23 s, respectively—both lower than the Transformer‑Regression baseline (MAE\,=\,60.53 s, RMSE\,=\,116.21 s).

\subsection{Error by predictive horizon}

Table~\ref{tab:hor} breaks the MAE down into six 5‑minute predictive horizon bins. Predictions in the 0-5 minutes bin correspond to events where the \textit{observed} arrival time is within 0 to 5 minutes of the \emph{snapshot}.  
As a result, the trends observed in the aggregate metrics can be more sharply analysed.

For Transformers, BC reduces MAE by \(21.2\%\) at 0–5 min to \(2.9\%\) at 25–30 min relative to Regression; DCIL further cuts the error by \(27.2\%\) and \(8\%\), respectively, delivering the lowest values in every horizon. 
These gains indicate that DCIL has an advantage over BC, most notably in longer horizon predictions, highlighting covariate shift mitigation. 

With MLP models, BC reduces MAE by \(23.4\%\) at 0–5 min to \(2.8\%\) at 25–30 min compared with Regression; DCIL brings the gains to \(26.3\%\) and \(4.3\%\), respectively, mirroring the Transformer’s pattern.

Using XGBoost models, BC reduces MAE by \(10.7\%\) at 0–5 min to \(1\%\) at 5-10 min, but increases it by \(2.5\%\) at 10-15 min to \(9.8\%\) at 25–30 min relative to Regression. It performs better for short-range predictions and worse for the rest. Within the simulation setting, XGBoost’s errors grow faster with horizon than the deep models’, indicating a tree-specific sensitivity to distribution shift rather than a limitation of the simulation framework. In particular, the XGBoost policy is not explicitly trained as a stochastic policy or calibrated probabilistic model, so treating its outputs as action probabilities can amplify compounding errors under simulation.

DCIL with the Transformer remains the overall winner, yielding the best MAE across all predictive horizons. 
Notably, MLP‑BC beats Transformer regression up until 20 minutes, while MLP-DCIL beats it all the way to 25 minutes, despite a much simpler model complexity.
Taken together with the aggregate results, these horizon‑wise findings confirm that simulation is more effective than regression for 30-minute delay prediction, and that DCIL effectively reduces the effect of covariate shift.

\subsection{Discussion}

As noted in Section~\ref{sec:relwork}, imitation learning is not limited to behavioural cloning. Generative Adversarial Imitation Learning (GAIL) is the standard non‑BC alternative. Yet all of our GAIL runs with Transformer models collapsed: despite using PPO and applying stability tricks such as label smoothing, policy BC pretraining, gradient-norm clipping and reward shaping—the policy converged to a single constant action prediction, producing high validation MAE. We suspect the discriminator’s reward couples token (train) contexts in the Transformer, creating noisy, non-local gradients on individual actions. Decoupling state–action pairs could stabilise learning, but would discard key inter-train dependencies. This failure illustrates the complexity of adversarial imitation learning and motivates simpler, more stable training schemes such as DCIL that still counteract covariate shift.

Experimental results outline the superiority of simulation for train delay prediction compared with regression for deep learning methods. Most notably, simpler neural networks trained with an imitation learning scheme have better results than a Transformer using regression despite having $\approx 14 \times$ fewer parameters and a less expressive architecture. This highlights that the training objective, rather than raw model capacity, is the primary lever for accuracy on this task.

Additionally, DCIL's advantage over BC is more pronounced for Transformers than for MLP. This hints that DCIL scales 
with model expressiveness: the richer inductive bias of self‑attention allows the Transformer to exploit the trajectory‑level gradients that DCIL provides, whereas the lower‑capacity MLP already harvests most of the benefit from simply matching expert actions step‑by‑step.

A natural next step is to test whether the simulation‑based objectives retain their edge when labelled data are scarce. Because DCIL supplies a trajectory‑level reward that is generated on‑the‑fly by the simulator, it can augment each real sequence with many synthetic roll‑outs, effectively multiplying the learning signal. We therefore hypothesise that, under progressive down‑sampling of the training set, the gap between DCIL and both BC and regression will widen—especially for the Transformer, whose larger capacity typically demands more data.

Table \ref{tab:hor} shows that the absolute MAE difference between Regression and BC narrows as the forecast horizon grows
(e.g.\ 6.5 s at 0–5 min vs. 2.6 s at 25–30 min). In contrast, the Regression–DCIL gap stays roughly constant ($\approx$ 8 s), indicating that DCIL controls horizon‑dependent drift more effectively than BC. Investigating whether the same trend holds for horizons longer than 30 minutes is an interesting direction for future work.

\subsection{Uncertainty Quantification}

\begin{figure}[ht]
    \centering
    \includegraphics[width=\linewidth]{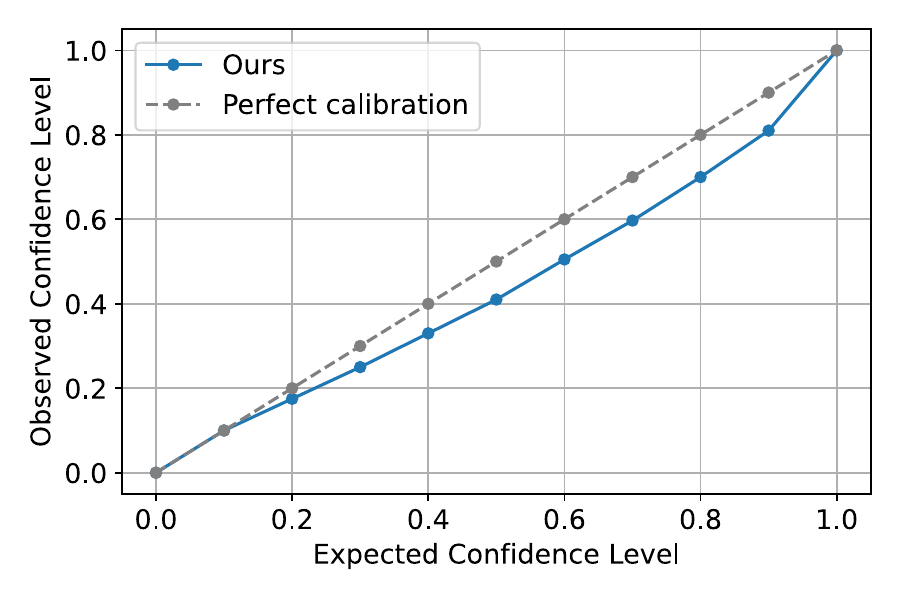}
    \caption{Calibration curve.}
\label{fig:calib}
\end{figure}

Following the procedure presented in \cite{kuleshov2018accurate}, we produce the calibration plot Fig.~\ref{fig:calib}; prediction intervals are formed from rollout percentiles, and their reliability is evaluated via percentile-based PIT calibration. The coverage curve lies consistently below the diagonal, indicating that the model is moderately overconfident: a nominal 80\% prediction interval, for instance, contains the ground‑truth delay only about 70\% of the time. We attribute this under-coverage to difficulty modelling extreme-delay events—rare but operationally important—which the ensemble rarely samples, causing mass to concentrate in the right-most bin and leaving earlier bins under-populated. We hypothesise that this may, in part, stem from inputs that only weakly encode the precursors of extreme delays.

\section{Conclusions and Future Work}\label{sec:conclusions}


In this paper, we framed probabilistic delay prediction as a sequential simulation task and trained an autoregressive policy via imitation learning using Behavioural Cloning and our novel approach, \textit{Drift-Corrected Imitation Learning} (DCIL), a self-supervised extension of DAgger that curbs the covariate shift that plagues Behavioural Cloning without relying on an external oracle or complex adversarial schemes.

Through extensive evaluation on large-scale real-world data from \textsc{Infrabel}, covering over three million train movements spanning three years, DCIL demonstrated superior performance compared with traditional regression methods and Behavioural Cloning on deep learning architectures for all predictive horizons.

Future work could refine the simulation by incorporating higher‑resolution representations of the railway network, leveraging GPS‑based data for added detail. A second line of inquiry involves investigating DCIL’s applicability to other imitation learning scenarios beyond delay prediction, assessing its generalisability and robustness across diverse domains. Furthermore, examining the theoretical and practical connections between DCIL and reinforcement learning—particularly model-based reinforcement learning—could provide valuable insights. Additionally, investigating the integration of DCIL within model-predictive control frameworks, specifically regarding performance and scalability over extended prediction horizons, presents a promising avenue for enhancing operational decision-making in complex, dynamic environments.

%% file: appendix.tex

\appendix

\section{Computing Infrastructure}

Data processing and experiments are run in a high performance cluster using Linux and Slurm.

\paragraph{Data processing} Data processing is conducted using 40$\times$ Intel(R) Xeon(R) Gold 6248 CPU @ 2.50GHz processors and 200Go of RAM totalling 15 hours.
\paragraph{Transformer} Transformer experiments are conducted using A100 gpus, 8$\times$ EPYC 7543 Milan AMD processors and 64Go of RAM totalling 2000 hours.
\paragraph{MLP/XGBoost} MLP and XGBoost experiments are conducted using V100 gpus, 10$\times$ Intel(R) Xeon(R) Gold 6248 CPU @ 2.50GHz processors and 40Go of RAM totalling 1500 hours.

\section{Hyper-parameter Search Protocol} 

Across all three model families—\textbf{Transformer}, \textbf{Multi-Layer Perceptron (MLP)}, and \textbf{XGBoost}—we perform a three-stage grid search. 
At each stage, we sweep the hyper-parameters listed in Tables~\ref{tab:transformer-sweep}--\ref{tab:mlp-sweep}--\ref{tab:xgb-sweep} over all Cartesian products, 
fixing all other settings to the best configuration from the previous stage. 
Validation uses mean absolute error (MAE) on the delay prediction targets, with a variance-aware criterion: 
if two candidates have similar MAE, the one with lower training variance is selected. 
For Transformers and MLPs, we apply early stopping on the validation MAE with patience equal to $0.25$ of the maximum epoch count 
(e.g., 20 epochs for an 80-epoch run). 

After tuning, the best configuration is retrained on the union of train and validation data 
and evaluated on the test split using ten random seeds (0–9), 
with all randomness controlled via PyTorch Lightning's global seeding.
In all tables, a dash (---) indicates that the field is not applicable to the corresponding method.

\paragraph{Transformer.}  The Transformer sweep supports \emph{Regression}, \emph{Behavioural Cloning (BC)} and \emph{Drift-Corrected Imitation Learning (DCIL)}.  All variants share the optimiser and architectural defaults listed at the top of Table~\ref{tab:transformer-sweep}.  Phase 1 explores model dimension, number of layers and learning rate, Phase 2 fine-tunes dropout, batch size and learning rate, and Phase 3 (DCIL only) searches over trajectory length, $\alpha$ and $\beta$.

\begin{table*}[p]
\centering
\caption{Default hyper-parameters and grid-search ranges for Transformer-based methods.}
\label{tab:transformer-sweep}
\begin{tabular}{llccc}
\hline
\textbf{Phase} & \textbf{Hyper-parameter} & \textbf{Regression} & \textbf{BC} & \textbf{DCIL} \\
\hline
\multicolumn{5}{l}{\textit{Defaults}}\\
 & Optimiser & \multicolumn{3}{c}{AdamW (default $\alpha,\;\beta$), weight decay 0.01}\\
 & Activation & \multicolumn{3}{c}{ReLU}\\
 & $d_{\mathrm{ff}}$ & \multicolumn{3}{c}{$4\,d_{\mathrm{model}}$}\\
 & Loss & L2 & Cross-entropy & Cross-entropy\\
 & Training epochs & 80 & 80 & 600\\
 & Batch size & 64 & 64 & 128\\
 & Heads $n_{\mathrm{head}}$ & \multicolumn{3}{c}{8}\\
 & Dropout & \multicolumn{3}{c}{0.2}\\
 & Replay buffer & --- & --- & 60,000\\
 & Synthetic samples/epoch & --- & --- & 20,000\\
 & Trajectory length & --- & --- & 10\\
  & $\alpha$ & --- & --- & 0.5\\
   & $\beta$ & --- & --- & 2\\
 
\hline
\multicolumn{5}{l}{\textit{Phase 1}}\\
 & $d_{\mathrm{model}}$ & \multicolumn{3}{c}{\{128, 256, 512, 1024\}}\\
 & Layers & \multicolumn{3}{c}{\{4, 6\}}\\
 & Learning rate & \multicolumn{3}{c}{\{1e-4, 5e-5, 1e-5\}}\\
\hline
\multicolumn{5}{l}{\textit{Phase 2}}\\
 & Batch size & \multicolumn{3}{c}{\{64, 128, 256\}}\\
 & Dropout & \multicolumn{3}{c}{\{0.05, 0.10, 0.20\}}\\
 & Learning rate & \{1e-4, 5e-5, 2e-5\} & \{3e-4, 1e-4, 5e-5\} & same as BC\\
\hline
\multicolumn{5}{l}{\textit{Phase 3 (DCIL only)}}\\
 & Trajectory length & --- & --- & \{5, 10, 15, 20\}\\
 & $\alpha$ & --- & --- & \{0.5, 0.8\}\\
 & $\beta$ & --- & --- & \{1, 2, 3, 4\}\\
\hline
\end{tabular}
\end{table*}

\paragraph{MLP.}  The MLP sweep supports \emph{Regression}, \emph{Behavioural Cloning (BC)} and \emph{Drift-Corrected Imitation Learning (DCIL)}.  All variants share the optimiser and architectural defaults listed at the top of Table~\ref{tab:mlp-sweep}.  Phase 1 explores hidden dimensions sizes and learning rate, Phase 2 fine-tunes batch size and learning rate, and Phase 3 (DCIL only) searches over trajectory length, $\alpha$ and $\beta$.

\begin{table*}[p]
\centering
\caption{Default hyper-parameters and grid-search ranges for MLP-based methods.}
\label{tab:mlp-sweep}
\begin{tabular}{llccc}
\hline
\textbf{Phase} & \textbf{Hyper-parameter} & \textbf{Regression} & \textbf{BC} & \textbf{DCIL} \\
\hline
\multicolumn{5}{l}{\textit{Defaults}}\\
 & Optimiser & \multicolumn{3}{c}{AdamW (default $\alpha,\;\beta$), weight decay 0.001}\\
 & Activation & \multicolumn{3}{c}{ReLU}\\
 & Loss & L2 & Cross-entropy & Cross-entropy\\
 & Training epochs & 100 & 160 & 1500\\
 & Batch size & \multicolumn{3}{c}{32}\\
 & Dropout & \multicolumn{3}{c}{0.0}\\
 & Replay buffer & --- & --- & 30,000\\
 & Synthetic samples/epoch & --- & --- & 10,000\\
  & Trajectory length & --- & --- & 10\\
  & $\alpha$ & --- & --- & 0.5\\
   & $\beta$ & --- & --- & 2\\
\hline
\multicolumn{5}{l}{\textit{Phase 1}}\\
 & Hidden Dims & \multicolumn{3}{c}{\{(64, 128, 256, 128, 64) to (256, 512, 1024, 2048, 1024, 512, 256)\} (8 configs)}\\
 & Learning rate & \multicolumn{3}{c}{\{1e-3, 5e-4, 1e-4\}}\\
\hline
\multicolumn{5}{l}{\textit{Phase 2}}\\
 & Batch size & \multicolumn{3}{c}{\{16 32 64 128 256\}}\\
 & Learning rate & \{3e-4, 1e-4, 5e-5, 3e-5\} & \{3e-3, 1e-3, 3e-4, 1e-4\} & same as Regression\\
\hline
\multicolumn{5}{l}{\textit{Phase 3 (DCIL only)}}\\
 & Trajectory length & --- & --- & \{5, 10, 15, 20\}\\
 & $\alpha$ & --- & --- & \{0.5, 0.8\}\\
 & $\beta$ & --- & --- & \{1, 2, 3, 4\}\\
\hline
\end{tabular}
\end{table*}

\paragraph{XGBoost.}  The XGBoost sweep supports \emph{Regression} and \emph{Behavioural Cloning (BC)}.  All variants share the optimiser and architectural defaults listed at the top of Table~\ref{tab:xgb-sweep}.  Phase 1 explores gamma, max depth, min child weight, subsample and colsample by tree and Phase 2 fine-tunes learning rate, number of estimators, reg $\alpha$ and reg $\lambda$.

\begin{table*}[p]
\centering
\caption{Default hyper-parameters and grid-search ranges for XGBoost-based methods.}
\label{tab:xgb-sweep}
\begin{tabular}{llcc}
\hline
\textbf{Phase} & \textbf{Hyper-parameter} & \textbf{Regression} & \textbf{BC} \\
\hline
\multicolumn{4}{l}{\textit{Defaults}}\\
 & Loss & L2 & Softprob\\
 & \# Estimators & \multicolumn{2}{c}{400} \\
 & Learning Rate & \multicolumn{2}{c}{0.1} \\
 & Reg $\alpha$ & \multicolumn{2}{c}{0} \\
& Reg $\lambda$ & \multicolumn{2}{c}{1} \\
 
\hline
\multicolumn{4}{l}{\textit{Phase 1}}\\
 & $\gamma$ & \multicolumn{2}{c}{\{0, 1, 5\}}\\
& Max Depth & \multicolumn{2}{c}{\{4, 6, 9, 13\}}\\
& Min Child Weight & \multicolumn{2}{c}{\{1, 5\}}\\
& Subsample & \multicolumn{2}{c}{\{0.6, 0.8, 1.0\}}\\
& Colsample by Tree & \multicolumn{2}{c}{\{0.5 0.8\}}\\
\hline
\multicolumn{4}{l}{\textit{Phase 2}}\\
& Learning Rate & \multicolumn{2}{c}{\{0.03, 0.04, 0.06, 0.07, 0.09, 0.1\}}\\
& \# Estimators & \multicolumn{2}{c}{\{200, 400, 800, 1000, 1600, 2000\}}\\
& Reg $\alpha$ & \multicolumn{2}{c}{\{0, 0.3, 1\}}\\
& Reg $\lambda$ & \multicolumn{2}{c}{\{0, 1, 5\}}\\
\hline
\end{tabular}
\end{table*}

\section{Final configurations}

The final configurations are given in Tables~{\ref{tab:transformer-best}-\ref{tab:mlp-best}-\ref{tab:xgb-best}}.

\begin{table*}[p]
\centering
\caption{Best hyper-parameters for Transformer models.}
\label{tab:transformer-best}
\begin{tabular}{llccc}
\hline
\textbf{Hyper-parameter} & \textbf{Regression} & \textbf{BC} & \textbf{DCIL} \\
\hline
 $d_{\mathrm{model}}$ & 512 & 512 & 512 \\
 Layers & 4 & 6 & 4 \\
 Learning rate & 5e-5 & 5e-5 & 1e-4 \\
 Batch size & 64 & 128 & 64 \\
 Dropout & 0.2 & 0.2 & 0.2 \\
 Trajectory length & --- & --- & 10\\
 $\alpha$ & --- & --- & 0.8\\
 $\beta$ & --- & --- & 2\\
\hline
\end{tabular}
\end{table*}

\begin{table*}[p]
\centering
\caption{Best hyper-parameters for MLP models.}
\label{tab:mlp-best}
\begin{tabular}{llccc}
\hline
\textbf{Hyper-parameter} & \textbf{Regression} & \textbf{BC} & \textbf{DCIL} \\
\hline
 Hidden dims & (512, 1024, 2048, 1024, 512) & (128, 256, 512, 1024, 512, 256, 128) & (256, 512, 1024, 512, 256) \\
 Learning rate & 1e-4 & 1e-3 & 5e-5 \\
 Batch size & 32 & 32 & 16 \\
 Trajectory length & --- & --- & 5\\
 $\alpha$ & --- & --- & 0.5\\
 $\beta$ & --- & --- & 1\\
\hline
\end{tabular}
\end{table*}

\begin{table*}[p]
\centering
\caption{Best hyper-parameters for XGBoost models.}
\label{tab:xgb-best}
\begin{tabular}{llcc}
\hline
\textbf{Hyper-parameter} & \textbf{Regression} & \textbf{BC} \\
\hline
$\gamma$ & 0 & 1 \\
Max Depth & 13 & 13 \\
Min Child Weight & 5.0 & 5.0 \\
Subsample  & 1.0 & 1.0 \\
Colsample by Tree & 0.8 & 0.8 \\
Learning Rate  & 0.03 & 0.04 \\
\# Estimators  & 2000 & 1600 \\
Reg $\alpha$ & 1.0 & 0.3 \\
Reg $\lambda$  & 5.0 & 1.0 \\ 
\hline
\end{tabular}
\end{table*}

\section*{Ethics Statement}

We use railway operational data released under a CC0 public-domain license. The dataset contains only non-personal operational information (train times, stations, delays) and no data about individual passengers or staff. Our use of the data therefore complies with the provider’s license and does not raise additional privacy concerns.

To ensure reproducibility and support further research, we release all scripts needed to reproduce our experiments; the corresponding GitHub link is provided in the Links section at the beginning of the paper.

\section*{Acknowledgements}

This work received financial support from SNCF through the research chair “AI and optimization for mobility” with École Polytechnique. This work was granted access to the HPC resources of IDRIS under the allocation AD011015656R1 made by GENCI. Finally, we thank Clément Mantoux and Mathis Le Bail for helpful discussions and feedback on this work.
